\documentclass[11pt]{article}

\usepackage[utf8]{inputenc}
\usepackage[T1]{fontenc}
\usepackage{lmodern}
\usepackage{microtype}
\usepackage{amsmath,amssymb,amsfonts}
\usepackage{graphicx}
\usepackage{booktabs}
\usepackage{multirow}
\usepackage{float}
\usepackage{subcaption}
\usepackage{xcolor}
\usepackage{url}
\usepackage{array}
\usepackage{tikz}
\usetikzlibrary{shapes.geometric, arrows.meta, positioning}
\usepackage{enumitem}
\usepackage{algorithm2e}
\usepackage{setspace}
\usepackage{abstract}
\usepackage{titlesec}
\usepackage[numbers]{natbib}
\usepackage[colorlinks=true,linkcolor=blue,citecolor=blue,urlcolor=blue]{hyperref}
\usepackage[a4paper,margin=1in]{geometry}

\titleformat{\section}{\large\bfseries}{\thesection}{1em}{}
\titleformat{\subsection}{\normalsize\bfseries}{\thesubsection}{1em}{}
\titleformat{\subsubsection}{\normalsize\itshape}{\thesubsubsection}{1em}{}

\title{\textbf{Confidence-Guided Diffusion Augmentation for Enhanced Bangla Compound Character Recognition}}

\author{
Md. Sultan Al Rayhan\\
Department of Computer Science and Engineering\\
East West University, Dhaka, Bangladesh\\
\texttt{salraihan444@yahoo.com}
}

\date{May 2026}

\begin{document}

\maketitle


\begin{abstract}
Recognition of handwritten Bangla compound characters remains a challenging problem due to complex character structures, large intra-class variation, and limited availability of high-quality annotated data. Existing Bangla handwritten character recognition systems often struggle to generalize across diverse writing styles, particularly for compound characters containing intricate ligatures and diacritical variations. In this work, we propose a confidence-guided diffusion augmentation framework for low-resolution Bangla compound character recognition. Our framework combines class-conditional diffusion modeling with classifier guidance to synthesize high-quality handwritten compound character samples. To further improve generation quality, we introduce Squeeze-and-Excitation enhanced residual blocks within the diffusion model's U-Net backbone. We additionally propose a confidence-based filtering mechanism where pre-trained classifiers act as quality gates to retain only highly class-consistent synthetic samples. The filtered synthetic images are fused with the original training data and used to retrain multiple classification architectures. Experiments conducted on the AIBangla compound character dataset demonstrate consistent performance improvements across ResNet50, DenseNet121, VGG16, and Vision Transformer architectures. Our best-performing model achieves 89.2\% classification accuracy, surpassing the previously published AIBangla benchmark by a substantial margin. The results demonstrate that quality-aware diffusion augmentation can effectively enhance handwritten character recognition performance in low-resource script domains.
\end{abstract}


\vspace{1em}
\hrule
\vspace{1.5em}

\section{Introduction}

Handwritten Character Recognition (HCR) plays a critical role in bridging the gap between handwritten human communication and machine-readable digital systems. By converting handwritten text into structured digital representations, HCR enables efficient archival, search, retrieval, and automated analysis of textual information. The field has witnessed substantial progress in recent years due to advances in deep learning and computer vision, particularly through convolutional neural networks and transformer-based architectures \cite{lecun2015deep}.

Despite major advances in optical character recognition for Latin scripts, handwritten recognition for low-resource Indic languages remains comparatively underexplored. Bangla (Bengali), one of the world's most widely spoken languages with over 240 million speakers, presents unique recognition challenges due to its complex script structure, cursive writing patterns, and extensive use of compound characters \cite{eberhard2023ethnologue}. Compound characters are formed through combinations of multiple consonants and often exhibit highly intricate visual patterns that vary significantly across handwriting styles.

The recognition of Bangla compound characters is substantially more difficult than isolated character recognition for several reasons. First, compound characters contain visually dense ligatures with subtle structural variations. Second, handwritten styles vary significantly across individuals, educational backgrounds, and geographic regions. Third, publicly available annotated datasets remain limited relative to large-scale datasets available for English or Chinese handwriting recognition. Finally, many existing systems rely on high-resolution inputs and computationally expensive training procedures that limit practical deployment in resource-constrained environments.

Recent progress in generative modeling, particularly diffusion models, has demonstrated remarkable capability in producing high-fidelity synthetic images \cite{ho2020denoising,dhariwal2021diffusion,rombach2022ldm}. Diffusion-based image synthesis has shown superior diversity and stability compared to earlier generative adversarial network approaches. However, the application of diffusion models to low-resource handwritten script augmentation remains relatively limited.

In this paper, we propose a confidence-guided diffusion augmentation framework for Bangla handwritten compound character recognition. Our framework integrates a class-conditional diffusion model with classifier guidance to generate high-quality synthetic compound character images. We further introduce Squeeze-and-Excitation enhanced residual blocks into the diffusion U-Net backbone to improve feature representation and generation quality. To prevent low-quality synthetic samples from degrading classifier performance, we introduce a confidence-based filtering mechanism that selectively retains only highly class-consistent synthetic samples.

The proposed framework operates on low-resolution 32\(\times\)32 grayscale images, making it computationally efficient while remaining suitable for real-world document analysis scenarios. Extensive experiments on the AIBangla compound character dataset demonstrate that the proposed framework consistently improves classification performance across both convolutional and transformer-based architectures.

The major contributions of this work are summarized as follows:

\begin{enumerate}[leftmargin=1.2cm]
    \item We propose a confidence-guided diffusion augmentation framework for Bangla compound character recognition.
    \item We introduce Squeeze-and-Excitation enhanced residual blocks within the diffusion model U-Net backbone to improve synthetic image quality.
    \item We develop a classifier-confidence-based filtering strategy to remove low-quality synthetic samples before retraining.
    \item We demonstrate substantial performance improvements across multiple classifier architectures on the AIBangla benchmark dataset.
    \item We establish new benchmark performance on the AIBangla compound character recognition task.
\end{enumerate}

\section{Related Work}

Research on Bangla handwritten character recognition and synthetic data generation has evolved significantly in recent years. This section briefly reviews prior work related to Bangla handwritten recognition, diffusion-based synthesis, and synthetic data quality control.

\subsection{Bangla Handwritten Character Recognition}

Research on Bangla handwritten character recognition has expanded significantly over the past decade. Early approaches relied heavily on handcrafted feature extraction methods such as Histogram of Oriented Gradients (HOG), Local Binary Patterns (LBP), and directional stroke features combined with classical machine learning classifiers including Support Vector Machines and Random Forests \cite{kibria2020bangla,sarkar2012cMATERdb}.

With the emergence of deep learning, convolutional neural networks rapidly became dominant for handwritten character recognition. Hasan et al. introduced the AIBangla dataset containing 249,911 handwritten Bangla compound character images across 171 classes and established baseline CNN results \cite{hasan2019aibangla}. Subsequent studies demonstrated improved performance through transfer learning and deeper architectures. Fardous et al. applied CNN-based models to Bangla compound characters and achieved strong recognition performance on the CMATERdb dataset \cite{fardous2019handwritten}. Hasan et al. further explored transfer learning with ResNet50 for Bangla handwritten character recognition \cite{hasan2020bengali}.

More recent work has investigated lightweight and attention-enhanced architectures. Khan et al. proposed a Squeeze-and-Excitation ResNeXt architecture for Bangla handwritten character recognition \cite{khan2022squeeze}. Hasan et al. introduced ComNet, incorporating EfficientNet-based feature extraction for compound character classification \cite{hasan2022comnet}. Ahmed et al. proposed a multilayer CNN framework with box detection for compound Bangla character recognition \cite{ahmed2023cnn}.

Although these methods demonstrate promising performance, most existing approaches rely exclusively on real training data and remain sensitive to class imbalance and limited handwriting diversity.

\subsection{Synthetic Data Generation for Handwritten Recognition}

Synthetic data augmentation has emerged as an effective strategy for improving handwritten recognition systems under limited data conditions. Early approaches relied primarily on geometric transformations, elastic distortions, and stroke perturbations \cite{simard2003best}. Generative Adversarial Networks (GANs) later enabled the synthesis of more realistic handwritten samples \cite{goodfellow2014generative,mirza2014conditional}.

For Bangla handwriting specifically, Nishat et al. explored Conditional GANs for generating isolated Bangla handwritten characters \cite{nishat2019synthetic}. However, GAN-based training often suffers from instability, mode collapse, and limited diversity.

Diffusion models have recently emerged as a more stable and scalable generative paradigm. Ho et al. introduced Denoising Diffusion Probabilistic Models (DDPMs), demonstrating state-of-the-art image synthesis quality \cite{ho2020denoising}. Dhariwal and Nichol further improved controllable image generation using classifier guidance \cite{dhariwal2021diffusion}. Song et al. proposed DDIMs for accelerated diffusion sampling \cite{song2021ddim}. Rombach et al. introduced latent diffusion models for efficient high-resolution image generation \cite{rombach2022ldm}.

Within Bangla handwriting research, Fuad et al. proposed Okkhor-Diffusion for generating isolated Bangla handwritten characters \cite{fuad2024okkhor}. However, compound character synthesis remains relatively unexplored due to the significantly larger and more complex class space.

\subsection{Synthetic Data Quality Filtering}

A major challenge in synthetic augmentation is ensuring the quality and class consistency of generated samples. Poor-quality synthetic images can negatively impact downstream classification performance. Several studies have explored selective augmentation and confidence-based filtering strategies \cite{xue2019selective}.

Inspired by these approaches, we incorporate classifier-confidence-based filtering into the augmentation pipeline. Unlike prior Bangla handwriting studies that directly use generated images without quality verification, our framework selectively retains only high-confidence synthetic samples, improving the reliability of the augmented dataset.

\section{Methodology}

This section presents the proposed confidence-guided diffusion augmentation framework for Bangla compound character recognition. We first describe the dataset and preprocessing pipeline, followed by the diffusion-based synthetic image generation framework, classifier-guided filtering mechanism, and downstream classification setup.

\subsection{Dataset}

We use the compound character subset of the AIBangla dataset \cite{hasan2019aibangla}, which contains 249,911 handwritten Bangla compound character images distributed across 171 classes. The dataset was collected from over 2,000 writers representing diverse demographic and educational backgrounds.

The dataset is divided into training, validation, and test splits using stratified sampling:

\begin{itemize}
    \item Training set: 199,928 images
    \item Validation set: 24,991 images
    \item Test set: 24,992 images
\end{itemize}

All images are converted to grayscale and resized to 32\(\times\)32 resolution.

\subsection{Framework Overview}

\begin{figure}[H]
\centering
\begin{tikzpicture}[
node distance=1.8cm,
every node/.style={font=\small},
process/.style={rectangle, rounded corners, minimum width=3.8cm, minimum height=1cm, text centered, draw=black},
arrow/.style={-{Latex[length=3mm]}, thick}
]

\node (data) [process] {Real AIBangla Dataset};
\node (diffusion) [process, below of=data] {Train Diffusion Model};
\node (generate) [process, below of=diffusion] {Generate Synthetic Images};
\node (filter) [process, below of=generate] {Confidence-Based Filtering};
\node (fusion) [process, below of=filter] {Fuse Real and Synthetic Data};
\node (train) [process, below of=fusion] {Retrain Classification Models};
\node (result) [process, below of=train] {Performance Evaluation};

\draw [arrow] (data) -- (diffusion);
\draw [arrow] (diffusion) -- (generate);
\draw [arrow] (generate) -- (filter);
\draw [arrow] (filter) -- (fusion);
\draw [arrow] (fusion) -- (train);
\draw [arrow] (train) -- (result);

\end{tikzpicture}
\caption{Workflow of the proposed confidence-guided diffusion augmentation framework.}
\label{fig:workflow}
\end{figure}
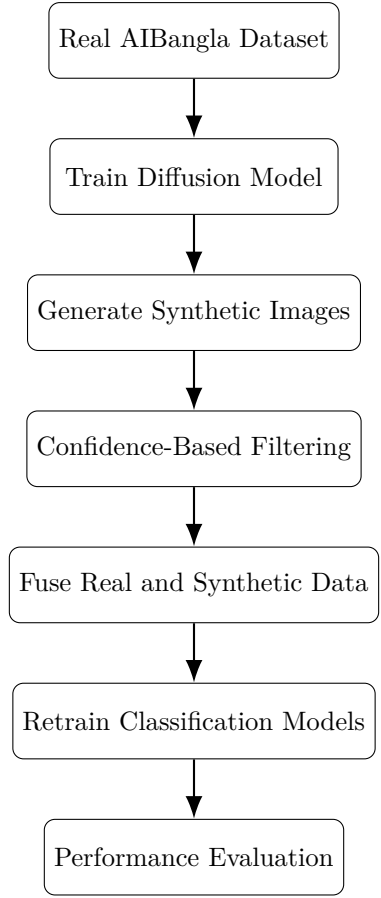

The overall workflow of the proposed framework is illustrated in Figure~\ref{fig:workflow}. The pipeline begins with real Bangla compound character images, followed by diffusion-based synthetic image generation, classifier-based quality filtering, dataset fusion, and final classifier retraining.

The proposed framework consists of four primary stages:

\begin{enumerate}[leftmargin=1.2cm]
    \item Train a class-conditional diffusion model with classifier guidance.
    \item Generate synthetic Bangla compound character images.
    \item Filter synthetic images using classifier confidence thresholds.
    \item Retrain classification architectures on fused real and synthetic datasets.
\end{enumerate}

\subsection{Diffusion Model Architecture}

The diffusion model employs a U-Net backbone enhanced with Squeeze-and-Excitation residual blocks. The architecture consists of encoder blocks, bottleneck attention modules, and decoder blocks connected through skip connections.

Key architectural components include:

\begin{itemize}
    \item Time embeddings using sinusoidal positional encoding.
    \item Learned class embeddings.
    \item SE-ResNet blocks for channel attention.
    \item Linear attention in encoder and decoder stages.
    \item Multi-head attention within the bottleneck.
\end{itemize}

The network predicts the injected noise \(\epsilon_\theta(x_t,t,y)\) at each diffusion timestep.

\subsection{Classifier Guidance Network}

An auxiliary classifier is trained on noisy images to guide the diffusion sampling process. The classifier shares structural similarities with the diffusion encoder and outputs class probabilities conditioned on noisy inputs.

During sampling, classifier gradients steer generated images toward target classes.

\subsection{Synthetic Image Filtering}

To improve augmentation quality, generated synthetic images are filtered using pre-trained classifiers. Only samples with confidence scores greater than or equal to 0.90 for their intended class are retained.

This process significantly reduces low-quality or ambiguous synthetic samples.

\subsection{Classifier Training}

We evaluate four classification architectures:

\begin{itemize}
    \item ResNet50
    \item DenseNet121
    \item VGG16
    \item Vision Transformer (ViT)
\end{itemize}

All models are trained using AdamW optimization with learning rate \(10^{-4}\), batch size 128, and early stopping.

\section{Experiments}

We conduct extensive experiments to evaluate the effectiveness of the proposed augmentation framework across multiple classification architectures. This section describes the experimental setup, implementation details, and evaluation metrics.

\subsection{Experimental Setup}

Experiments are conducted using NVIDIA Tesla P100 GPUs on the Kaggle cloud platform. All implementations are developed using PyTorch.

\subsection{Evaluation Metrics}

We evaluate classification performance using:

\begin{itemize}
    \item Accuracy
    \item Precision
    \item Recall
    \item F1-score
    \item Fr\'echet Inception Distance (FID)
\end{itemize}

Lower FID values indicate greater similarity between generated and real image distributions.

\section{Results and Discussion}

This section presents quantitative and qualitative evaluations of the proposed framework. We analyze diffusion quality, filtering effectiveness, and classification improvements.

\subsection{Diffusion Quality}

The proposed diffusion model achieves an FID score of 24.63 after 300 epochs of training. The generated images capture significant stylistic diversity while preserving class-specific structure.

\subsection{Filtering Performance}

Table~\ref{tab:fid} presents FID scores after classifier-based filtering.

\begin{table}[H]
\centering
\caption{FID scores for filtered synthetic datasets.}
\label{tab:fid}
\begin{tabular}{lcc}
\toprule
Dataset & Images Retained & FID \\
\midrule
Unfiltered & 57,285 & 24.63 \\
VGG16 Filtered & 47,485 & 24.39 \\
ResNet50 Filtered & 29,885 & 24.17 \\
ViT Filtered & 17,743 & 24.04 \\
DenseNet Filtered & 33,915 & 24.09 \\
\bottomrule
\end{tabular}
\end{table}

Filtering consistently improves FID across all architectures.

\subsection{Classification Performance}

\begin{table}[H]
\centering
\caption{Baseline classifier performance.}
\label{tab:baseline}
\begin{tabular}{lcccc}
\toprule
Model & Loss & Accuracy & Precision & Recall \\
\midrule
ResNet50 & 0.786 & 0.787 & 0.7874 & 0.7856 \\
DenseNet121 & 0.772 & 0.788 & 0.7879 & 0.7876 \\
VGG16 & 0.560 & 0.882 & 0.8391 & 0.8345 \\
ViT & 1.265 & 0.670 & 0.6689 & 0.6692 \\
\bottomrule
\end{tabular}
\end{table}

\begin{table}[H]
\centering
\caption{Performance after retraining with filtered synthetic data.}
\label{tab:retrained}
\begin{tabular}{lcccc}
\toprule
Model & Loss & Accuracy & Precision & Recall \\
\midrule
ResNet50 & 0.710 & 0.806 & 0.8066 & 0.8058 \\
DenseNet121 & 0.734 & 0.797 & 0.7967 & 0.7962 \\
VGG16 & 0.515 & 0.892 & 0.8470 & 0.8441 \\
ViT & 1.136 & 0.696 & 0.6938 & 0.6941 \\
\bottomrule
\end{tabular}
\end{table}

All architectures demonstrate improved performance after augmentation. VGG16 achieves the best overall performance with 89.2\% accuracy.

\subsection{Discussion}

The proposed framework demonstrates that diffusion-based synthetic augmentation can significantly improve Bangla handwritten compound character recognition. The improvements are consistent across both convolutional and transformer-based architectures, suggesting strong generalization.

The confidence-based filtering mechanism plays an important role in maintaining augmentation quality. Filtering removes low-confidence synthetic samples and improves the overall quality of the augmented dataset.

Operating at low image resolution further demonstrates the practicality of the framework for deployment in real-world document analysis systems.

\section{Limitations}

Although the proposed framework achieves strong performance improvements, several limitations remain. First, diffusion sampling remains computationally expensive despite operating at low image resolution. Second, FID may not fully capture semantic quality for handwritten character synthesis. Third, the framework currently focuses exclusively on isolated compound character recognition rather than full handwritten word recognition.

Future work may explore latent diffusion models, faster sampling strategies, and sequence-level handwritten text recognition.

\section{Conclusion}

This paper presented a confidence-guided diffusion augmentation framework for Bangla handwritten compound character recognition. By combining class-conditional diffusion synthesis, classifier guidance, SE-enhanced U-Net architecture, and confidence-based filtering, the proposed framework generates high-quality synthetic handwritten samples that substantially improve downstream classification performance.

Extensive experiments on the AIBangla dataset demonstrate consistent improvements across multiple classifier architectures. The proposed framework establishes new benchmark performance while operating efficiently at low image resolution.

The results highlight the potential of quality-aware generative augmentation for low-resource handwritten recognition tasks and provide a promising direction for future research in Indic script document analysis.

\section*{Acknowledgements}

The author thanks the Department of Computer Science and Engineering at East West University for institutional support. The author also acknowledges Kaggle for providing cloud-based computational resources used in this research.


\end{document}